\RequirePackage[l2tabu, orthodox]{nag}

\documentclass[utf8,biblatex,norunningheads,english]{lnimodified}

\usepackage{booktabs}
\usepackage[utf8]{inputenc}

\iffalse
\AtEveryBibitem{%
  \ifentrytype{article}{%
  }{%
    \clearfield{doi}%
    \clearfield{issn}%
    \clearfield{url}%
    \clearfield{urldate}%
  }%
  \ifentrytype{inproceedings}{%
  }{%
    \clearfield{doi}%
    \clearfield{issn}%
    \clearfield{url}%
    \clearfield{urldate}%
  }%
}
\fi

\begin{document}

\title[Evolutionary Computation and the Voynich Manuscript]{Can Evolutionary Computation Help us to Crib the Voynich Manuscript ?}
\author[Daniel Devatman Hromada]
{Daniel Devatman Hromada\footnote{Einstein Center Digital Future and Faculty of Design of Berlin University of the Arts (dh at udk-berlin dot de)}}
\maketitle

\begin{abstract}

Voynich Manuscript is a corpus of unknown origin written down in unique graphemic system and potentially representing phonic values of unknown or potentially even extinct language. Departing from the postulate that the manuscript is not a hoax but rather encodes authentic contents, our article presents an evolutionary algorithm which aims to find the most optimal mapping between voynichian glyphs and candidate phonemic values. 

Core component of the decoding algorithm is a process of maximization of a fitness function which aims to find most optimal set of substitution rules allowing to  transcribe the part of the manuscript - which we call the Calendar - into lists of \textbf{feminine names}. This leads to sets of character subsitution rules which allow us to consistently transcribe dozens among three hundred calendar tokens into feminine names: a result far surpassing both ``popular'' as well as ``state of the art'' tentatives to crack the manuscript. What's more, by using name lists stemming from different languages as potential cribs, our ``adaptive'' method can also be useful in identification of the language in which the manuscript is written. 

As far as we can currently tell, results of our experiments indicate that the Calendar part of the manuscript contains names from \textbf{baltoslavic, balkanic or hebrew language strata}. Two further indications are also given: primo, highest fitness values were obtained when the crib list contains names with specific infixes at token's penultimate position as is the case, for example, for slavic \textbf{feminine diminutives} (i.e. names ending with -ka and not -a). In the most successful scenario, 240 characters contained in 35 distinct Voynichese tokens were successfully transcribed. Secundo, in case of crib stemming from Hebrew language, whole adaptation process converges to significantly better fitness values when transcribing voynichian tokens whose order of individual characters have been reversed, and when lists feminine and not masculine names are used as the crib.

\end{abstract}


\newpage

\section{Introduction}
Voynich Manuscript (VM) undoubtably counts among the most famous unresolved enigmas of the medieval period. On approximately 240 vellum pages currently stored as manuscript (MS) 408 in Yale University's Beinecke Rare Book and Manuscript Library, VM contains many images apparently related to botanics, astronomy (or astrology) and bathing. Written aside, above and below these images are bulks of sequences of glyphs. All this is certain.

Also certain seems to be the fact that in 1912, VM was re-discovered by a polish book-dealer Wilfrid Voynich in a large palace near Rome called Villa Mandragone. Alongside the VM itself, Voynich also found the correspondence - dating from 1666 - between Collegio Romano scholar Athanasius Kircher and the contemporary rector of Charles University in Prague, Johannes Marcus Marci. Other attested documents - e.g. a letter from 1639 sent to Kircher by a Prague alchemist Georg Baresch - also indicate that during the first half of 17th century, VM was to be found in Prague. The very same correspondence also indicates that VM was acquired by famous patron of arts, sciences and alchemy, Emperor Rudolf II. \footnote{Savants which passed through Rudolf's court included Johannes Kepler, Tycho deBrahe or Giordanno Bruno. The last one is known to have sold a certain book to the emperor for 600 ducats.}

Asides this, one more fact can be stated with certainty: the vellum of VM was carbon-dated to the early 15h century \citep{hodgins2014forsenic}.

\subsection{Pre-digital tentatives}
Already during the preinformatic era of first half of 20th century had dozens, if not hundreds, men of distinction invested non-negligeable time of their life into tentatives to decipher the ``voynichese'' script. 

Being highly popular in their time, many such tentatives - like that of Newbold who claimed to ``prove'' that VM was encoded by Roger Bacon by means of 6-step anagrammatic cipher \citep{newbold}, or that of Strong \citep{strong1945anthony} who claimed VM to be a 16th-century equivalent of the Kinsey Report'' - may seem to be, when looked upon through the prism of computer science, somewhat irrational \footnote{Note, for example, Strong's ``translation'' of one VM passage: ``\emph{When the contents of the veins rip, the child comes slyly from the mother issuing with leg-stance skewed and bent while the arms, bend at the elbow, are knotted like the legs of a crawfish.}'' \cite{strong1945anthony} Note also that such translation was a product of man who was ``a highly respected medical scientist in the field of cancer research at Yale University'' \citep{d1978voynich}.}

C.f. \citep{d1978voynich} for a overview of other 20th-century ``manual'' tentatives which resulted in VM-decipherement claims. After description of these tentatives and and after presentation of informationally very rich introduction to both VM and its historical context, d'Imperio adopts a sceptical stance towards all scholars who associated VM's origin with the personage of Roger Bacon\footnote{"I feel, in sum, that Bacon was not a man who would have produced a work such as the Voynich manuscript...I can far more easily imagine a small society perhaps in Germany or Eastern Europe \citep[51]{d1978voynich}"}. 

In spite of sceptic who she was, d'Imperio hadn't a priori disqualified a set of hypotheses that the language in which the VM was ultimately written was latin or medieval English. And such, indeed, was the majority of hypotheses which gained prominence all along 20th century.\footnote{Note that such pro-english and pro-latin bias can be easily explained not by the properties of VM itself, but by the simple fact that first batches of VM's copies were primarily distributed and popularized among anglosaxxon scholars of medieval philosophy, classical philology or occidental history}.

\subsection{Post-digital tentatives}
First tentatives to use machines to crack the VM date back to prehistory of informatic era. Thus, already during 2nd world war did the cryptologist William F. Friedman invited his colleagues to form ``extracurricular'' VM study group - programming IBM computers for sorting and tabelation of VM data was one among the tasks. Two decades later - and already in position of a first chief cryptologist of the nascent National Security Agency - Friedman had formed the 2nd study group. Again without ultimate success.

One member of Friedman's 2nd Study Group After was Prescott Currier whose computer-driven analysis led him to conclusion that VM in fact encodes two ``statistically distinct'' \citep{currier19701976} languages. What's more, Currier seems to have been the first scholar who facilitated the exchange and processing of Voynich manuscript by proposing a transliteration\footnote{In this article we distinguish transliteration and transcription. Transliteration is a bijective mapping from one graphemic system into another (e.g. VM glyphs is transliterated into ASCII's EVA subset). Transcription is a potentially non-bijective mapping between symbols one one side and sound- or meaning- carrying units on the other.}  of voynichese glyphs into standard ASCII characters. This had been the predecessor of the European Voynich Alphabet (EVA) \citep{landini1998well} which had become a de facto standard when it comes to mapping of VM glyphs upon the set of discrete symbols. 

Canonization of EVA combined with dissemination of VM's copies through Internet have allowed more and more researchers to transcribe the sequence of glyhps on the manuscript into ASCII EVA sequences. Is is thanks to laborious transcription work of people like Rene Zandberger, Jorge Stolfi or Takeshi Takahashi that verification or falsification of VM-related hypotheses can be nowadays in great extent automatized. 

For example, Stolfi's analyses of frequencies of occurence of different characters in different contexts has indicated that majority of Voynichese words seems to implement a sort of tripartite crust-core-mantle (or prefix, infix, suffix) morphology. Later study has indicated that the presence of such morphological regularities could be explained as an output of a mechanical device called Cadran grill \cite{rugg2004elegant}. The ``hoax hypothesis'' is also supported by the study \citep{schinner2007voynich} which suggested that ``the text has been generated by a stochastic process rather than by encoding or encryption of language". Pointing in the similar direction, the analysis also concludes that ``glyph groups in the VM are not used as words".

On the other hand, a methodology based on ``first-order statistics of word properties in a text, from the topology of complex networks representing texts, and from intermittency concepts where text is treated as a time series'' presented in \citep{amancio2013probing} lead its authors to conclusion that VM ``is mostly compatible with natural languages and incompatible with random texts". Simply stated, the way how diverse ``words'' are distributed among different sections of VM indicates that these words carry certain semantics. And this indicates that VM, or at least certain parts of it, are not a hoax.


\subsection{Our position}
Results of \citep{amancio2013probing} had made us adopt the conjecture ``VM is not a hoax'' as a sort of a fundamental hypothesis accepted \emph{a priori}. Surely, as far as we stand, it could not be excluded that VM is a work of an abnormal person, of somebody who suffered severe schizophrenia or was chronically obsessed by internal glossolalia \citep{kennedy2005voynich}. Nor can it be excluded that the manuscript does not encode full-fledged utterances but rather lists of indices, sequences or proper names of spirits-which-are-to-be-summoned or sutra-like formulas compressed in a sort of private pidgin or a sociolect. But given VM's ingenuity and given the effort which the author had to invest into the conception of the manuscript and given a sort of ``elegant simplicity'' which seems to permeate the manuscript, we have felt, since our very first contact with the manuscript, a sort of obligation to interpret its contents as meaningful.

That is, as having the capability of denoting the objects outside of the manuscript itself. As being endowed with the faculty of reference to the world \citep{frege1994sinn} which we, 21st century interpretators, still inhabit hundred years after VM's most plausible date of conception. 

It is with such bias in mind that our attention was focused upon a certain regularity which we have later decided to call ``the primary mapping".

\subsection{Primary Mapping}
\emph{Condition sine qua non} of any act of decipherement is a discovery of rules which allow to transform initially meaningless cipher into meaningful information. In most trivial case, such decipherement is facilitated by a sort of Rosetta Stone \citep{champollion1822observations} which the decipherer already has at his disposition. Since both the ciphertext as well as the plaintext (also called ``the crib") are explicitely given by the Rosetta Stone, discovery of the mapping between the two is usually quite straightforward.

The problem with VM is, of course, that it seems not to contain any explicit key which could help us to decipher its glyphs. Thus, the only source of information which could potentially help us to establish reference between VM's glyphs and the external world are VM's drawings. One such drawing present atop of folio f84r is shown on Fig. 1.

Figure 1 displays twelve women bathing in eight compartments of a pool. Bathing women is a very common motive present in VM and there seems to be nothing peculiar about it. The fact that word-like sequences are written above heads of these women is also trivial. 

\begin{figure}
	\includegraphics[width=\textwidth]{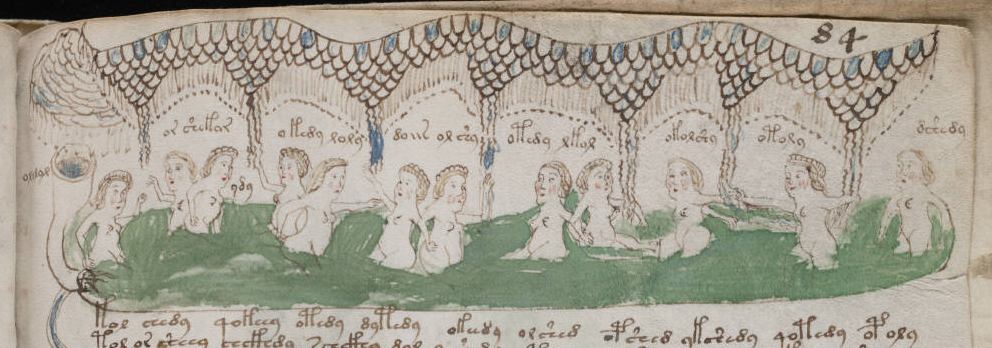}
	\caption{Drawing from fiolio f84r containing the primary mapping.}
\end{figure}

One can, however, observe one regularity which seems to be interesting. That is, in case two women bath in the same compartement, the compartement contains two word-like sequences. If one woman bathes in the compartement, there is only one word-like sequence which is written above her head. 

One figure - one word, two figures - two words. This principle is stringently followed and can be seen on other folios as well. What is more, the words themselves are sometimes similiar but they are not the same. Such trivial observations lead to trivial conclusion: these word-like sequences are labels. 

And since these names are juxtaposed to feminine figures, it seems reasonable to postulate that these labels are, in fact, feminine names. This is the primary mapping.


\subsection{Three Conjectures}

Method which shall be described in following sections can be considered as valid only under assumption that following conjectures are valid:
\begin{enumerate}
	\item ``the primary mapping conjecture'' : voynichese words asides feminine figures are feminine names
	\item ``diachronic stability of proper names'' : proper names are less prone to diachronic change than other language units
	\item ``occam's razor'' : instead of containing a sophisticated esoteric cipher, VM simply transmits a text written in an unknown script
\end{enumerate}

Further reasons why we consider ``the primary mapping conjecture'' as valid shall be given alongside our discussions of ``the Calendar". When it comes to conjecture postulating the ``diachronic stability of proper names", we could potentially refer to certain cognitive peculiarities or how human mind tends to treat proper names \citep{imai2001learning}. Or focus the attention of the reader to the fact that for practically every human speaker, one's own name undoubtably belongs among the most frequent and most important tokens which one hears or utters during whole life - this can result in a sort of stability against linguistic change and allow the name to cross the centuries with higher probability than words of lesser importance and frequency. 

But instead of pursuing the debate in such a direction, let's just point out that successful decoding of Mycenian Linear script B (\citep{ventris1953evidence} would be much more difficult if certain toponyms like Amnisos, Knossos or Pylos haven't succeeded to carry their phonetic skeleton through aeons of time.

At last but not least, the ``occam's razor conjecture'' simply explicitates the belief that a reasonable scientist should not opt to explain VM in terms of annagrams and opaque hermeneutic procedures if similar - or even more plausible - results can be attained when approaching VM as it was a simple substitution cipher.

\section{Method}	
The core of our method is an optimization algorithm which looks for such a candidate transcription alphabet $A_x$ which, when applied upon the list of word types occurent in VM's Calendar section yields an output list whose members should be ideally present in another list, called the Crib. The optimization is done by an evolutionary strategy - an individual chromosome encode a candidate transcription alphabet and a fitness function is given as a sum of lengths of all tokens which were successfully transcribed from Calendar to a specified Crib. 

\subsection{Calendar}
Six among twelve words present on Figure 1. occur only on folio f84r. Six others occur on other folios as well, and five of these six words occur also as labels near feminine figures displayed on 12 folios of the section commonly known as ``Zodiac". It is like this that our attention was focused from the limited corpus of ``primary mapping'' towards more exhaustive corpus contained in the Zodiac.

Every page of Zodiac displays multiple concentric circles filled with feminine figures. Attributes of these figures differ - some hold torches, some do not, some are bathing, some are not - but one pattern is fairly regular. Asides every woman there is a star and asides every star, there is a word.

While some authors postulate that these words are names of stars or names of days, we postulate that these words are simply feminine names\footnote{It cannot be excluded, however, that they all this at once. Note, for example, that in many central european countries, it is still a fairly common practice to attribute specific names to specific days in a year, i.e. ``meniny". }. From Takahashi's transliterations of twelve folios of the Zodiac we extract 290 tokens which instantiate 264 distinct word types.

To evit possible terminological confusion, we shall denote this list of 264 labels\footnote{Available at http://wizzion.com/thesis/simulation0/calendar.uniq} with the term Calendar. Hence, Zodiac is the term to refer to folios f70v2 - f73v, while Calendar is simply a list of 264 labels. Total length of this 264 labels is 2045 letters. These characters are chosen from 19-symbol ($|A_{cipher}|=19$) subset of the EVA transliteration alphabet.

\subsection{Cribbing}
Cribbing is a method by means of which a hypothesis, that the Calendar contains lists of feminine names, can potentially lead to decipherment of the manuscript. For if the Calendar is indeed such a list, then one could use lists of existing and attested feminine names as hypothetic target ``cribs". 

In cryptanalytic terms, an intuition that the Calendar contains feminine names makes it possible to perform a sort of known-plaintext attack (KPA). We say ``a sort of", because in case of VM are the ``cribs'' upon which we shall aim to map the Calendar, not known with 100\% certainity. Hence, it is maybe more reasonable to understand the cribbing procedure as the plausible-plaintext attack (PPA).

This beings said, we label as ``cribbing'' a symbol-substituting procedure $P_{cribbing}$ which replaces symbols contained in the cipher (i.e. in the Calendar) with symbols contained in the plaintext. Hence, not only cipher but also plaintext are inputs of the cribbing procedure.

Every act of execution of $P_{cribbing}$ can be followed an act of evaluation of usefulness $P_{cribbing}$ in regards to its inputs. The ideal procedure would result in a perfect match between the rewritten cipher and the plaintext, i.e.

\[P_{cribbing}(cipher) == plaintext\]

On the other hand, a completely failed $P_{cribbing}$ results in two corpora which do not have anything in common.

And between two extremes of the spectrum, between ``the ideal'' and ``the completely failed", one can place multitudes other procedures, some closer to the ideal than the others. This makes place for optimization.

\subsection{Optimization}
All experiments described in the next section of this article implement an evolutionary computation algorithm which strongly inspired by the architecture of canonic genetic algorithm (CGA, P+46) \cite{holland:ga2,rudolph1994convergence}. Hence, initial population is randomly generated and the fitness-proportionate (i.e. ``roullette wheel", P+42) selection is used as the main selection operator. But contrary to CGAs, our optimization technique does not implement a classical single-point crossover but rather a sort of ``discrete crossover'' which takes place only in case that parent individuals have different alelles of a specific gene.

\definecolor{Green}{RGB}{0,255,0}
\begin{lstlisting}[language=Perl,float,floatplacement=t,caption=Discrete cross-over]
#discrete crossover
my $child_genome;
my $i=0;
for (@mother_genome) {
	if ($_ ne $father_genome[$i]) {
		rand > 0.5 ? ($child.=$mother_genome[$i]) : ($child.=$father_genome[$i]);
	} else {
		$child_genome.=$mother_genome[$i];
	}
	$i++;
}

\end{lstlisting}

Another reason why our solution can be considered to be more similar to evolutionary strategies \citep{rechenberg:es} than to CGAs is related to the fact that it does not encode individuals as binary vector (P+48). Instead, \emph{every individual represents a candidate monoalphabetic substitution cipher} application of which could, ideally, transform the Calendar into a crib. More formally: given that cipher is written in symbols of the alphabet $A_{cipher}$ and given that the crib is written in symbols of the alphabet $A_{crib}$, then each individual chromosome will have length of $|A_{crib}|$ genes and every individual gene could encode one among $|A_{cipher}|$ values. 

Size of the search space is therefore $|A_{cipher}|^|{A_{crib}}|$. Search for optima in this space is governed by a fitness function:

\[F_{P_{cribbing}}=\sum_{w \in cipher \land P_{cribbing}(w) \in crib}{length(w)}\]

where $w$ is a word type occurent in the cipher (i.e. in the Calendar) and which, after being rewritten by $P_{cribbing}$ also matches a token in the input crib. Given that the expression $length(w)$ simply denotes $w$'s character length, the fitness function of the candidate transcription procedure $P_{cribbing}$ is thus nothing else than the sum of character lengths of all distinct labels contained in the Calendar which $P_{cribbing}$ successfully maps onto the feminine names contained in the input crib.

\definecolor{Green}{RGB}{0,255,0}
\begin{lstlisting}[language=Perl,float,floatplacement=t,caption=Cipher2Dictionary adaptation fitness function]
#Fitness Function
my $text=$calendar;
my $old = "acdefghiklmnopqrsty";
my %translit;
@translit{split //, $old} = split //, $individual;
$text =~ s/(.)/defined($translit{$1}) ? $translit{$1} : $1/eg; #core transcription of calendar content
my %matched;
for (split/\n/,$text) {
	my $token=$_;
	if (exists $crib{$token}) {
		@antitranslit{split //, $individual} = split //, $old;
		$token =~ s/(.)/defined($antitranslit{$1}) ? $antitranslit{$1} : $1/eg;
		my $t=$token;
		$matched{$t}=1;
	}
}
for (keys %matched) {
	$Fitness[$i]+=length $_;
}

\end{lstlisting}

\section{Experiments}
Within the scope of this article, we present results of two sets of experiments which essentially differed in the choice of a name-containing cribs. 

Other input values (e.g. Takahashi's transliteration of the Calendar used as the cipher) and evolutionary parameters (total population size = 5000, elite population size = 5, gene mutation probability \textless 0.001) were kept constant between all experiments and sub-experiments. Each experiment consisted of ten distinct runs. Each run was terminated after 200 generations.

\subsection{Slavic crib}
What we label as ``slavic crib'' is a plaintext list of feminine names which we had compiled from multiple sources publicly available on the Internet. Principal sources of names were websites of western slavic origin. This choice was motivated by following reasons:
\begin{enumerate}
	\item The oldest more or less certain trace of VM's trajectory points to the city of Prague - the center of western slavic culture.
	\item Ortography of western slavic languages relatively faithfully represent the pronounciation. That is, there are relatively few digraphs (e.g. a bigram ``ch'' which denotes a voiced velar fricative). Hene, the distance between the graphemic and the phonemic representations is not so huge as in case of english or french.
	\item Slavic languages have rich but regular affective and diminutive morphology which is often used when addressing or denoting beloved persons by their first name.
\end{enumerate}

The third reason is worth to be introduced somewhat further: in both slavic and western slavic languages, a simple infixing of the unvoiced velar occlusive ``k'' before the terminal vowel ``a'' of a feminine names leads to creation of a diminutive form of such a name (e.g. $alena \to alenka, helena \to helenka $ etc.) The fact that this morphological rule is used both by western as well as eastern slavs indicates that the rule itself can be quite old, date to \emph{common slavic} or even \emph{pre-slavic} periods and hence, was quite probably in action already in the period when VM was written. 

For the purpose of this article, let's just note that application of the substitution:
\[a\$ \to ka/\]
allowed us to significantly increase the extent of the ``slavic crib". Thus, we have obtained a list a of 13815 distinct word types which are in quite close relation to phonetic representation of feminine names used in europe and beyond\footnote{Slavic crib is publicly available at http://wizzion.com/thesis/simulation0/slavic\_extended.crib}. The alphabet of this crib comprises of 38 symbols, hence there exists $19^{39}$ possible ways how symbols of the Calendar could be replaced by symbols of this crib.
	
Figure 2. shows the process of convergence from populations of randomly generated chromosomes towards more optimal states. In case of runs averaged in the ``SUBSTITUTON'' curve, the procedure $P_{cribbing}$ consisted in simple mapping of the Calendar onto the crib by means of a substitution cipher specified in the chromosome. But in case of runs averaged in the ``REVERSAL + SUBSTITUTION'' curve, whole process was initiated by the reversal of order of characters present within individual tokens of the Calendar (e.g. $okedy \to ydeko, otedy \to ydeto$ etc.)
Let's now look at contens of individuals which were ``identified'' by the optimization method.

More concrete illustrations can also turn out to be quite illuminating. Hence, if the most elite individual of run 1 (i.e. the one with fitness 197) is as a means of substitution of EVA characters contained in the Calendar, one will see appearance of names like ALENA, ALETHE, ANNA, ATENKA, HANKA, HELENA, LENA etc. And when the last one (i.e. the one with fitness 240 is used), the resulting list shall contain tokens like AELLA, ALANA, ALINA, ANKA, ANISSA, ARIANNKA, ELLINA, IANKA, ILIJA, INNA, LILIJA, LILIKA, LINA, MILANA, MILINA, RANKA, RINA, TINA etc. 

This being said, the observation that all reversal-implementing runs have converged to genomes which:
\begin{enumerate}
	\item transcribe e in EVA as nasal n
	\item transcribe k in EVA as velar k
	\item transcribe t in EVA as nasal n
	\item transcribe y in EVA as vowel a
	\item transcribe a in EVA as vowel (80\% times as ``i", 10\% as ``e", 10\% as ``o")
	\item transcribe l in EVA as either a liquid consonant (80\% ``l", 10\% ``r") or ``m'' (10\%)
\end{enumerate}
...could also be of certain use and importance.

\begin{figure}
	\includegraphics[width=0.77\textwidth]{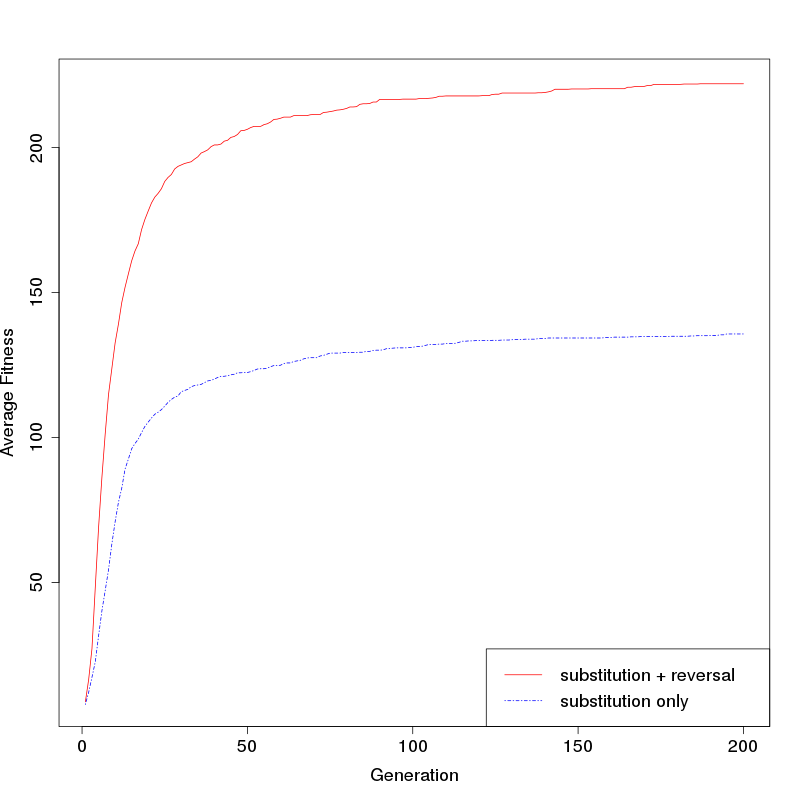}
	\caption{Evolution of individuals adapting label in the Calendar to names listed in the slavic crib.}
\end{figure}
\renewcommand{\tabcolsep}{1pt}
\begin{table}[h]
	\centering
	\begin{tabular}{c|ccccccccccccccccccc}
		Fitness & \\
		\toprule
197 & e&s&t&n&h&k&a&h&k&l&h&t&a&k&a&m&e&n&a    \\
230 & i&k&t&n&s&k&n&h&k&l&z&t&a&j&s&m&i&n&a    \\
224 & i&c&t&n&v&k&/&g&k&l&m&b&a&j&/&r&i&n&a    \\
227 & i& &t&n&p&a&f&l&k&l&m&e&a&n&k&r&i&n&a    \\
240 & i&k&t&n&a&k&f&l&k&l&m&e&a&j&g&r&i&n&a    \\
226 & i& &l&n&h&o& &l&k&r&g&e&a&n&a&m&i&n&a    \\
208 & i&q&g&n&x&k&d&e&k&l&m&x&a&j&x&r&i&n&a    \\
239 & i&k&t&n&d&o&l&l&k&l&f&e&a&k&i&m&i&n&a    \\
191 & o&t&l&n&t&n&n&r&k&m&z&b&a&n&h&r&e&n&a    \\
240 & i&s&t&n&s&k&n&l&k&l&m&e&a&j&I&r&i&n&a  \\
\bottomrule
EVA & a&c&d&e&f&g&h&i&k&l&m&n&o&p&q&r&s&t&y \\
	\bottomrule
	\end{tabular}
	\caption{Fittest chromosomes which map reversed tokens in the Calendar onto names of the slavic crib}
\end{table}

\subsection{Hebrew crib}
At this point, a sceptical mind could start to object that what our algorithm adapt to is in fact not the Calendar, but the statistical properties of the crib. And in case of such a long and sometimes somewhat artificial list like $Crib_{slavic}$, such an objection would be in great extent justified. For the adaptive tendencies of our evolutionary strategy are indeed so strong that it would indeed find a way to partially adapt the calendar to a crib which is long enough\footnote{\label{footnote9}This has been, indeed, shown by multiple micro-experiments which we do not report here due to the lack of space. No matter whether we use cribs as absurd as list of modern american names or enochian of John Dee and Edward Kelly, we could always observe a sort of adaptation marked by the increase of fitness. But it was never so salient as in case of $Crib_{slavic}$ or $Crib_{hebrew}$.}

\begin{figure}
	\includegraphics[width=0.77\textwidth]{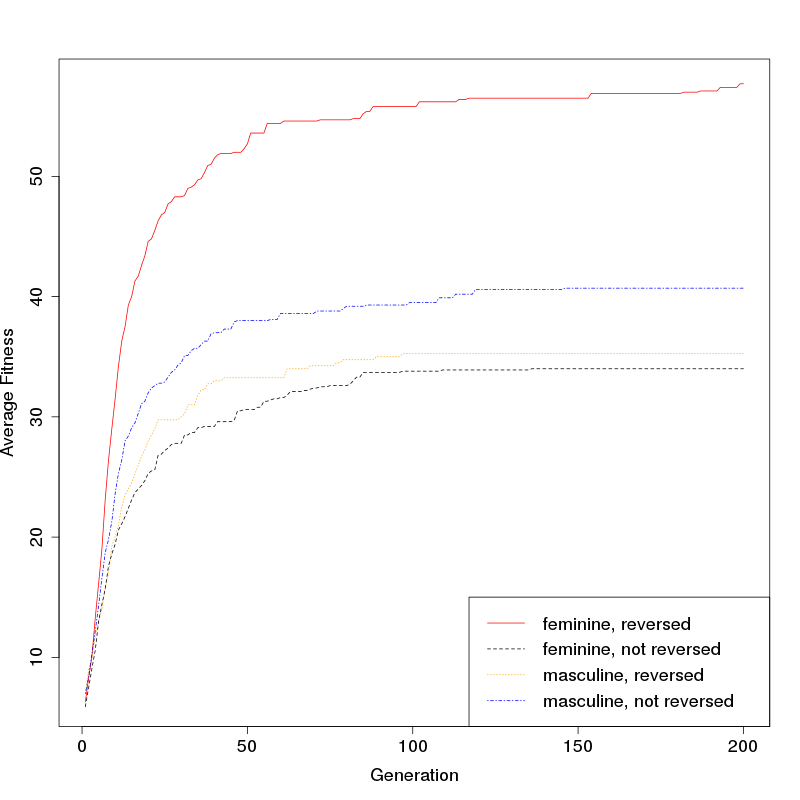}
	\caption{Evolution of individuals adapting label in the Calendar to names listed in the hebrew cribs.}
\end{figure}

For this reason, we have decided to target our second experiment not at the biggest possible crib but rather at the oldest possible crib. And given that our first experiment has indicated that it seems to be more plausible to interpret labels in the Calendar as if they were written in reverse, id est from right to left, our interest was gradually attracted by Hebrew language\footnote{Other reasons why we decided to focus on Hebrew include: important presence of Jewish diaspora in Prague of Rudolph the 2nd (c.f. the story of rabbi Loew and the Golem of Prague); ritual bathing of jewish women known as mikveh; usage of VM-ressembling triplicated forms (e.g. amen, amen, amen) in talmudic texts; attested existence of so-called Knaanic language which seems to be principially a czech language written in hebrew script et caetera et caetera.}. This lead us to two lists of names:
\begin{itemize}
	\item $Crib_{hebrew-men}$ contains 555 masculin names\footnote{http://wizzion.com/thesis/simulation0/jewish\_men}
	\item $Crib_{hebrew-women}$ contains 283 feminine names\footnote{http://wizzion.com/thesis/simulation0/jewish\_women}
\end{itemize}
both lists were extracted from the website finejudaica.com/pages/hebrew\_names.htm and were chosen because they did not contain any diacritics and hence transcribing hebrew names in a similiar way as they had been transcribed millenia ago.

Figure 3 displays the summary of all runs which aimed to transcribe the Calendar with hebrew names. As may be seen, the whole system converged to highest fitness values when $Crib_{hebrew-women}$ was used in concordance with reversal of order of characters. Difference results of these batch of runs and other results of other batches is statistically significant (p-value $<$ 7e-10).
.
%
%

The highest attained fitness value was was attained by the cribbing procedure which first reverses the order of characters whose EVA representations are subsequently substituted by a  following chromosome:

\includegraphics[width=\textwidth]{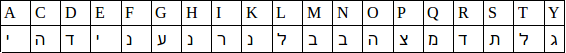}

This chromosome transcribes the voynichese Calendar labels \emph{okam, otainy, otey, oty, otaly, okaly, oky, okyd, ched, otald, orara, otal, salal and opalg} to feminine hebrew names 

	\includegraphics[width=\textwidth]{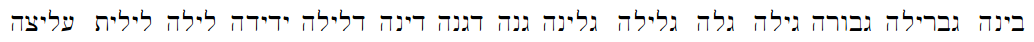}

(i.e. Bina, Gabriela, Ghila, Gala, Galila, Galina, Gina, Degana, Diyna, Deliyla, Yedidya, Lila, Lilit and Alica).

Worth mentioning are also some other fenomena related to these transcriptions. One can observe, for example, that the label ``otaly'' - translated as Galina - is also present on folios f33v, f34r or f46v which all contain drawings of torch-like plants. This is encouraging because the word ``galina'' is not only a hebrew name, but also a substantive meaning ``torch". Similary, the word ``lilit'' is not only a name but also means  ``of the night". This word supposedly translates the voynichese token ``salal'' which is very rare - asides the Calendar it occurs only on purely textual folio f58v and on a folio f67v2 which, surprise!, may well depict circadian rhytms of sunrise, sunset, day and night. 

Or it could be pointed out kind that the huge majority of occurences of voynichese trigram ``oky'' (potentially denoting the name ``gina'' which also means ``garden") is to be observed on herbal folios. Or the distribution of instances of ``okam'' (transcripted as ``bina'' which means ``intelligence and wisdom"\footnote{Note that ``bina'' is one among highest sephirots located at north-western corner of kabbalistic tree of life. In this context it is worth noting that only partially readable EVA group ``...kam'' occurs as a third word near the north-western ``rosette'' of folio 85v2. Such considerations, however, bring us too far.}  could, and potentially should, be taken into consideration. Or maybe not.


\section{Conclusion}

In 2013, BBC Online had anounced ``Breakthrough over 600-year-old mystery manuscript". The breakthrough was to be effectuated by Stephan Bax who, in his article, describes the process of decipherement as follows:

\citep{bax2014proposed}{The process can be compared to doing a crossword puzzle: at first we might doubt one possible answer in the crossword, but gradually, as we solve other words around it which serve to confirm letters we have already placed, we gradually gain more confidence in our first answer until eventually we are confident of the solution as a whole.}

What Bax does not add, unfortunately, is that the voynich crossword puzzle is so big that anyone who looks at it close enough can find in it small islands of order, local optima where few characters seem to fit the global pattern. Thus, even if Bax had succeeded, as he states, in ``identification of a set of proper names in the Voynich text, giving a total of ten words made up of fourteen of the Voynich symbols and clusters", this would mean nothing else than that he had identified a locally optimal transcription alphabet. 

In this article, we have presented two experiments employing two different lists of feminine names. Both experiments have indicated that if labels in the Zodiac encode feminine names, then these have been originally written from right to left \footnote{Note, however, that this does not necessarily imply that the scribe of VM (him|her)self had written the manuscript in right-to-left fashion. For example, in case (s)he was just reproducing an older source which (s)he didn't understand, his|her hand could trace movements from left to right while the very orignal had been written from right to left}. The first experiment led to identification of multiple substitution alphabets which allow to map 240 EVA letters, contained in 40 distinct words present in the Calendar, onto 35 feminine-name-ressembling sequences enumerated among 13815 items of $Crib_{Slavic}$. Results of second experiment indicate that if ever the Calendar contains lists of hebrew names, then these names would be more probably feminine rather than masculine.

This is, as far as we can currently say, all that could be potentially offered as an answer to the question \emph{Can Evolutionary Computation Help us to Crib the Voynich Manuscript?}. Everything else is - without help coming from experts in other disciplines - just a speculation.

\printbibliography

@preamble{ " \newcommand{\noop}[1]{} " }

@inproceedings{hodgins2014forsenic,
	  title={Forensic investigations of the Voynich MS},
	    author={Hodgins, G},
	      booktitle={Voynich 100 Conference www. voynich. nu/mon2012/index. html. Accessed},
	        volume={4},
		  year={2014}
}

@article{bax2014proposed,
	  title={A proposed partial decoding of the Voynich script},
	    author={Bax, Stephen},
	      journal={University of Bedfordshire, http://stephenbax. net/wp-content/uploads/2014/01/Voynich-a-provisionalpartial-decoding-BAX. pdf},
	        year={2014}
}

@article{landini1998well,
	  title={A well-kept secret of mediaeval science: The Voynich manuscript},
	    author={Landini, Gabriel and Zandbergen, Rene},
	      journal={Aesculapius},
	        volume={18},
		  pages={77--82},
		    year={1998}
}

@article{rugg2004elegant,
	  title={An elegant hoax? a possible solution to the Voynich manuscript},
	    author={Rugg, Gordon},
	      journal={Cryptologia},
	        volume={28},
		  number={1},
		    pages={31--46},
		      year={2004},
		        publisher={Taylor \& Francis}
}

@techreport{d1978voynich,
	  title={The Voynich manuscript: an elegant enigma},
	    author={d'Imperio, Mary E},
	      year={1978},
	        institution={DTIC Document}
}

@article{schinner2007voynich,
	  title={The Voynich manuscript: evidence of the hoax hypothesis},
	    author={Schinner, Andreas},
	      journal={Cryptologia},
	        volume={31},
		  number={2},
		    pages={95--107},
		      year={2007},
		        publisher={Taylor \& Francis}
}

@article{amancio2013probing,
	  title={Probing the statistical properties of unknown texts: application to the Voynich Manuscript},
	    author={Amancio, Diego R and Altmann, Eduardo G and Rybski, Diego and Oliveira Jr, Osvaldo N and Costa, Luciano da F},
	      journal={PloS one},
	        volume={8},
		  number={7},
		    pages={e67310},
		      year={2013},
		        publisher={Public Library of Science}
}

@book{kennedy2005voynich,
	  title={The Voynich manuscript: the unsolved riddle of an extraordinary book which has defied interpretation for centuries},
	    author={Kennedy, Gerry and Churchill, Rob},
	      year={2005},
	        publisher={Orion Publishing Company}
}

@article{ventris1953evidence,
	  title={Evidence for Greek dialect in the Mycenaean archives},
	    author={Ventris, Michael and Chadwick, John},
	      journal={The Journal of Hellenic Studies},
	        volume={73},
		  pages={84--103},
		    year={1953},
		      publisher={Cambridge Univ Press}
}

@article{holland:ga2,
          title={Genetic algorithms},
            author={Holland, John H},
              journal={Scientific american},
                volume={267},
                  number={1},
                    pages={66--72},
                      year={1992}
}

@article{rudolph1994convergence,
          title={Convergence analysis of canonical genetic algorithms},
            author={Rudolph, G{\"u}nter},
              journal={Neural Networks, IEEE Transactions on},
                volume={5},
                  number={1},
                    pages={96--101},
                      year={1994},
                        publisher={IEEE}
}

@phdthesis{rechenberg:es,
          title={Evolutionsstrategie: Optimierung technischer Systeme nach Prinzipien der biologischen Evolution. Dr.-Ing},
            author={Inno Rechenberg},
              year={1971},
                school={Thesis, Technical University of Berlin, Department of Process Engineering}
}

@article{frege1994sinn,
	  title={{\"U}ber sinn und bedeutung},
	    author={Frege, Gottlob},
	      journal={Wittgenstein Studien},
	        volume={1},
		  number={1},
		    year={1994}
}

@article{imai2001learning,
	  title={Learning proper nouns and common nouns without clues from syntax},
	    author={Imai, Mutsumi and Haryu, Etsuko},
	      journal={Child development},
	        volume={72},
		  number={3},
		    pages={787--802},
		      year={2001},
		        publisher={Wiley Online Library}
}

@article{strong1945anthony,
	  title={Anthony Askham, the author of the Voynich Manuscript},
	    author={Strong, Leonell C},
	      journal={Science},
	        volume={101},
		  number={2633},
		    pages={608--609},
		      year={1945},
		        publisher={American Association for the Advancement of Science}
}

@article{currier19701976,
	  title={1976." Voynich MS. Transcription Alphabet; Plans for Computer studies; Transcribed Text of Herbal A and B Material; Notes and Observations."},
	    author={Currier, Prescott},
	      journal={Unpublished communications to John H. Tiltman and M. D’Imperio, Damariscotta, Maine},
	        year={1970}
}

@book{champollion1822observations,
	  title={Observations sur l'obelisque Egyptien de l'Ile de Philae},
	    author={Champollion, Jean Fran{\c{c}}ois},
	      year={1822}
}

@book{newbold,
	  title={Cipher of Roger Bacon},
	    author={Newbold, William Romaine},
	      year={1928},
}
\end{document}